\documentclass[conference]{IEEEtran}

\usepackage{times}

\usepackage[numbers]{natbib}
\usepackage{multicol}
\usepackage[bookmarks=true]{hyperref}

\usepackage{graphicx} 
\usepackage{amsmath,amssymb}
\usepackage{nth}
\usepackage{color} 
\definecolor{mygreen}{RGB}{28,172,0} 
\definecolor{mylilas}{RGB}{170,55,241}
\usepackage{makecell}
\usepackage[utf8]{inputenc}
\usepackage{glossaries}
\usepackage{subcaption}
\usepackage[export]{adjustbox}
\usepackage{multirow}
\usepackage{svg}
\usepackage{wrapfig}
\usepackage{mathtools}

\usepackage[linesnumbered,ruled,vlined]{algorithm2e}

\SetCommentSty{mycommfont}
\usepackage{xcolor}

\def\HiLi{\leavevmode\rlap{\hbox to \hsize{\color{yellow!50}\leaders\hrule height .8\baselineskip depth .5ex\hfill}}}

\DeclareMathOperator*{\argmin}{arg\,min}

\usepackage{lipsum}
\usepackage{float}
\makeatletter
\def\BState{\State\hskip-\ALG@thistlm}
\makeatother
\newcommand{\norm}[1] {||#1||}
\usepackage{tabularx,ragged2e,booktabs,caption}

\newcolumntype{C}[1]{>{\Centering}m{#1}}

\newcommand{\normal}{\mathcal{N}}

\usepackage{flexisym}
\newcommand{\BEAS}{\begin{eqnarray*}}
\newcommand{\EEAS}{\end{eqnarray*}}
\newcommand{\BEA}{\begin{eqnarray}}
\newcommand{\EEA}{\end{eqnarray}}
\newcommand{\BEQ}{\begin{equation}}
\newcommand{\EEQ}{\end{equation}}
\newcommand{\BIT}{\begin{itemize}}
\newcommand{\EIT}{\end{itemize}}

\newcommand{\reals}{\mathbb{R}}

\newcommand{\diag}{\mathop{\textbf{diag}}}

   {\begin{list}{}{%
    \setlength{\rightmargin}{0\linewidth}%
    \setlength{\leftmargin}{.05\linewidth}}%
    \sffamily\small
    \item[]{\setlength{\parskip}{0ex}\hrulefill\par%
    \nopagebreak{}}}%
   {{\setlength{\parskip}{-1ex}\nopagebreak\par\hrulefill} \end{list}}

\graphicspath{{media/}}
\usepackage{mydefs}

\begin{document}

\title{\methodabv{}: Producing Diverse Plausible Pose Estimates from Contact and Free Space Data}



\author{\authorblockN{Sheng Zhong, Nima Fazeli, Dmitry Berenson}
\authorblockA{
University of Michigan, 
Ann Arbor, MI 48109\\
Email: \{zhsh, nfz, dmitryb\}@umich.edu}
}


%

\maketitle

\begin{abstract}
This paper proposes a novel method for estimating the set of plausible poses of a rigid object from a set of points with volumetric information, such as whether each point is in free space or on the surface of the object. In particular, we study how pose can be estimated
from force and tactile data arising from contact. 
Using data derived from contact is challenging because it is inherently less information-dense than visual data, and thus the pose estimation problem is severely under-constrained when there are few contacts. 
Rather than attempting to estimate the true pose of the object, which is not tractable without a large number of contacts, we seek to estimate a plausible set of poses which obey the constraints imposed by the sensor data. Existing methods struggle to estimate this set because they are either designed for single pose estimates or require informative priors to be effective. Our approach to this problem, Constrained pose Hypothesis Set Elimination (\textbf{\methodabv{}}), has three key attributes: 1) It considers volumetric information, which allows us to account for known free space; 2) It uses a novel differentiable volumetric cost function to take advantage of powerful gradient-based optimization tools; and 3) It uses methods from the Quality Diversity (QD) optimization literature to produce a diverse set of high-quality poses. To our knowledge, QD methods have not been used previously for pose registration. We also show how to update our plausible pose estimates online as more data is gathered by the robot. 
Our experiments suggest that \methodabv{} shows large performance improvements over several baseline methods for both simulated and real-world data. 
\footnote{This work was supported in part by Toyota Research Institute, the Office of Naval Research Grant N00014-21-1-2118 and NSF grants IIS-1750489, IIS-2113401, and  IIS-2220876. For code, see \href{https://github.com/UM-ARM-Lab/chsel}{\tt\small https://github.com/UM-ARM-Lab/chsel}}
\end{abstract}

\IEEEpeerreviewmaketitle

\section{Introduction}

\begin{figure}[ht!]
    \centering
    \includegraphics[width=\linewidth]{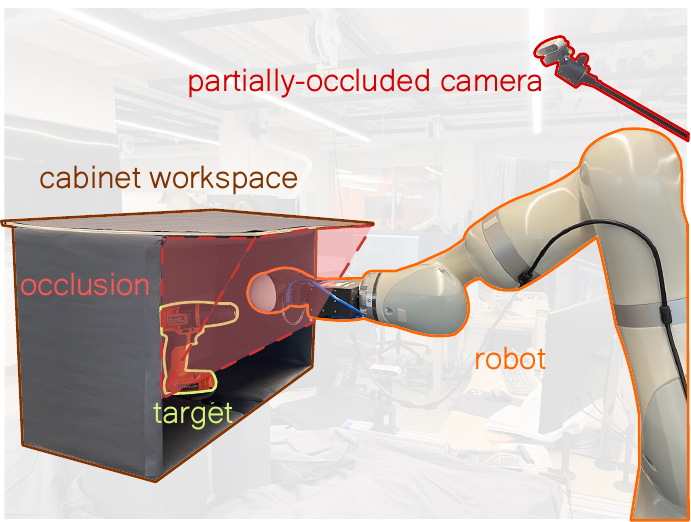}
    \includegraphics[width=0.43\linewidth]{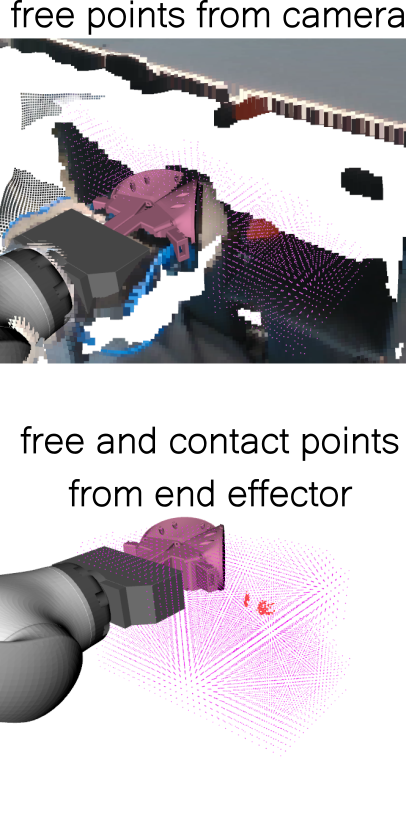}
    \includegraphics[width=0.55\linewidth]{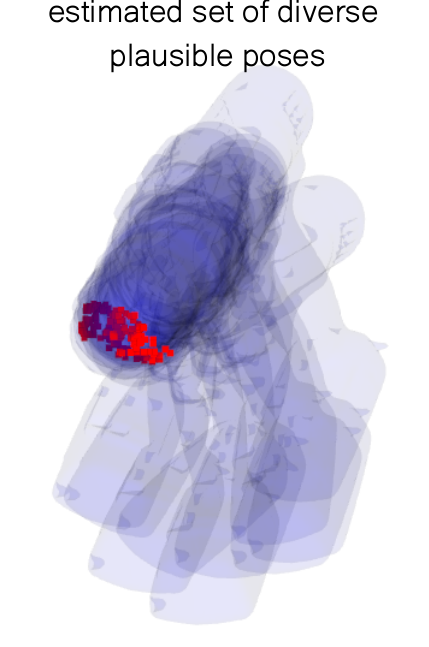}
    \caption{Top: Set up of a real-world probing experiment where the goal is to estimate the drill's pose. Bottom, left: input to \methodabv{}, made of known free points in pink (from the camera and swept robot volume) and known surface points in red (from contact). Bottom, right: \methodabv{} uses these points and the object model to estimate a diverse set of plausible poses.}
    \label{fig:teaser}
    
\vspace{-20pt}
\end{figure}

Pose registration---the process of estimating the pose of a given rigid object from sensor data, is a fundamental problem in robotics, as it is necessary for manipulation and reasoning. Much research has been done in estimating object pose from visual data, especially laser-range data ~\cite{collet2009object}~\cite{cheng2018registration}. However, a clear view of the object may not always be available (e.g. an object in a cupboard, as in Fig. \ref{fig:teaser}, or grocery bag) or the material properties of the object may make it difficult to perceive visually (e.g. transparency).

Partial occlusion in manipulation tasks motivates for rummaging, and researchers have investigated the use of tactile and force feedback for pose registration~\cite{sipos2022simultaneous}~\cite{dikhale2022visuotactile}. However, the nature of this data is quite different from the point-clouds produced by laser-scanners. While point-cloud data is information-dense (e.g. many points on the surface of the object), tactile and force data arising from contact contain much less information (e.g. one contact point per motion) in addition to being time-consuming to collect. This lack of information can be partially mitigated by assuming that a contact sensor moves along the surface of the object \cite{suresh2022midastouch,driess2017active}. However, creating controllers that can do this without moving the object is challenging.

In the context of pose registration problems, the lack of informative data results in a lack of constraints on the set of plausible poses of the object. In such cases, producing an accurate estimate of the true pose is very unlikely, and it is more useful to estimate the set of plausible poses. 
Especially towards the beginning of contact-based tasks,
uncertainty in the object pose is high due to insufficient data, sensor noise, and inherent object symmetries. Characterizing this uncertainty, such as in the form of a  set of plausible poses, is useful for object recognition~\cite{xu2022tandem3d}, active perception~\cite{eidenberger2010active}, and simultaneous localization and mapping (SLAM)~\cite{arras2003feature}~\cite{fu2021multi}. 

The most common methods for pose registration are based on the Iterative Closest Point (ICP) algorithm~\cite{segal2009generalized}~\cite{wang2017survey}, which outputs a single pose estimate for a given initial pose. These methods can be effective for point-cloud data, but producing a set of estimates from random initialization does not yield good coverage of the set of plausible poses for contact data. Bayesian
methods that aim to capture the full distribution of plausible object poses use approximation techniques such as Markov Chain Monte Carlo (MCMC) and variational inference~\cite{maken2021stein}. However, such variational methods depend heavily on informative priors, which we do not assume are available.

To overcome the above limitations, we present Constrained pose Hypothesis Set Elimination (\textbf{\methodabv{}}), which has three key attributes: First, we go beyond only considering points on the surface of the object, considering volumetric information instead (similar to~\citet{slavcheva2016sdf} and \citet{haugo2020iterative}). This allows us to infer more data (and thus more constraints on the pose) from robot motion. For example, when a robot moves into contact with an object, we observe contact points, as well as all the free space the robot traversed before and during contact. Note that this representation can also include free space and object surface points observed by a visual sensor. 

Second, to take advantage of powerful gradient-based optimization tools, we construct a differentiable cost function that can be used to efficiently optimize a given pose based on volumetric information. Finally, and most importantly, to estimate a diverse set of poses simultaneously, we adapt methods from the Quality Diversity (QD) optimization literature. To our knowledge, this work is the first application of QD methods to the problem of pose registration. QD methods explicitly optimize for a set of solutions which are both diverse and high-quality, making them a natural choice for pose registration problems that seek to capture the set of plausible poses. We also show how to update our set of estimates online as more data is gathered by the robot. 

Our experiments suggest that \methodabv{ }has large performance improvements over several baseline methods for both simulated and real-world contact data. Additionally, we compare against alternatives and show that our cost function is a good QD objective.
We also show that real-world visual data can be incorporated seamlessly into our cost function while demonstrating similar performance improvements.

\section{Related Work}
While our work is the first to use known free space to produce a diverse set of pose registrations, prior work has been done separately in using free space in registration and diverse set (also known as multi-hypothesis) registration.
Geometric registration has been extensively studied in robotics and computer vision (see~\citet{tam2012registration} for an overview). In particular, the distinction between free space and surface points can be framed as point semantics or features, and methods such as the 3D Normal Distribution Transform (3D NDT)~\cite{magnusson2007scan} and its continuous generalization Continuous Visual Odometry (CVO)~\cite{zhang2021new} have been designed with them in mind. We compare against CVO as a baseline. \citet{haugo2020iterative} considers free space explicitly, filling it with balls via the medial axis transformation. They then formulate a cost penalizing object-ball penetration while requiring points to lie on the surface. This is a baseline in our experiments.

Specific to SE(2) pose estimation in planar contact problems, the Manifold Particle Filter~\cite{koval2015pose} exploits a robot's contact manifold to estimate an pose. However, it struggles to scale to full SE(3) pose estimation as it is expected to require exponentially more particles. 

Deep learning based methods such as SegICP~\cite{wong2017segicp} and MHPE~\cite{fu2021multi} have been developed to produce a plausible set of pose estimates. However, they can only use points from the object surface and require relatively dense information.

Related to registration is the problem of object reconstruction. SDF-2-SDF~\cite{slavcheva2016sdf} minimizes the difference between pairs of signed distance fields (SDFs).
They construct an SDF using observed RGBD images and match it against the target SDF. In cases where a dense view of surface points is not available, such as when the camera is occluded or if the sensing is performed via contact, the constructed SDF will be invalid.  Similar to them, we directly work with the target object's SDF, but importantly do not assign SDF values to observed free points. Instead, we only require that known free points be outside the surface (SDF 0-level set).

Diversity in registration has mainly been explored as characterizing the pose uncertainty. ~\citet{censi2007accurate} provides a closed form estimate for ICP based methods, but require that the initial point-correspondences are correct and that the minimization procedure does not get caught in local minima. This is unlikely to be valid with partial information, and does not utilize known free space. ~\citet{buch2017prediction} produces high quality uncertainty estimates through MCMC simulation of a depth camera, which is computationally expensive while being restricted in input modality.~\citet{maken2021stein} performs Stein Variational Gradient Descent (SVGD) on a differentiable formulation of the ICP objective. They approximate the distribution of poses by running ICP from different starts, which in general is not the distribution of poses consistent with the data. We compare against SVGD optimization as a baseline.

Generating diverse sets of high quality solutions has been explored explicitly in recent research on evolutionary optimization techniques. In particular, Quality Diversity (QD)~\cite{pugh2016quality} techniques such as MAP-Elites~\cite{mouret2015illuminating} and CMA-MEGA~\cite{fontaine2021differentiable} have been developed to optimize objectives while enforcing diversity in some aspect of the solutions. We leverage QD optimization methods with our proposed differentiable cost function to estimate a set of plausible diverse transforms.
\section{Problem Statement}

For a target object, we have its precomputed object frame signed distance function (SDF) derived from its 3D model, $\sdf: \reals^3 \rightarrow \reals$ , and are given a set of points $\xall = \{(\x_1,\s_1),...,(\x_\numpoints, \s_\numpoints)\}$ with known world positions $\x_\n \in \reals^3$ and semantics $\s_\n$ (described below). $\xall$ is produced from sensor data. Object registration is the problem of finding transforms $\T \in \poseset$ that satisfy constraints imposed by $\xall$. Let $\T^*$ be the true object transform, then the semantics are
\begin{align}
\s_\n = \begin{cases} \free & \text{implies $\sdf(\T^*\x_\n) > 0$} \\
\occupied & \text{implies $\sdf(\T^*\x_\n) < 0$} \\
\kv_\n & \text{implies $\sdf(\T^*\x_\n) = \kv_\n$} \end{cases}
\end{align}

We quantify the degree to which the constraints of $\xall$ are satisfied by using a cost function (lower is better) $\totalcost(\xall, \T) = \sum_{n=1}^{N} \cost(\T\x_\n, \s_\n)$ where $\cmax \gg 0$ and

\begin{align}\label{eq:gtcost}
\cost(\x, \s) = \begin{cases} 
\cmax \indicator(\sdf(\x) \leq 0) & \text{if $\s = \free$} \\
\cmax \indicator(\sdf(\x) \geq 0) & \text{if $\s = \occupied$} \\
|\kv - \sdf(\x)| & \text{else}
\end{cases}
\end{align}

$\indicator$ is the indicator function that evaluates to 1 if the argument is true and otherwise evaluates to 0.
We then define the plausible set $\plausibleset_\suboptimality = \{\T \ | \ \totalcost(\xall, \T) - \totalcost(\xall, \T^*) < \suboptimality \}$ where $\suboptimality > 0$ is the degree of violation we allow in considering the constraints satisfied. Our goal is to produce a hypothesis transform set $\estimateset$ that covers $\plausibleset_\suboptimality$. To quantify how well we cover this set, we use the Plausible Diversity metric~\cite{saund2021diverse}~\cite{saund2022clasp}:
\begin{align}
\coverage &= \frac{1}{|\plausibleset_\suboptimality|} \sum_{\T \in \plausibleset_\suboptimality} \min_{\estT \in \estimateset} \tdist(\T, \estT) & \text{coverage} \\
\plausibility &= \frac{1}{|\estimateset|} \sum_{\estT \in \estimateset} \min_{\T \in \plausibleset_\suboptimality} \tdist(\T, \estT) & \text{plausibility} \\
\pd &= \coverage + \plausibility & \text{plausible diversity}
\end{align}
where $\tdist$ is a distance function on transforms, such as Chamfer Distance between the resulting transformed objects. This metric penalizes $\estimateset$ if 1) it does not include a transform that is close to each transform in the plausible set (a lack of coverage); or 2) includes transforms that are far from any transform in the plausible set (such transforms are implausible).

\begin{figure*}[ht]
\includegraphics[width=\linewidth]{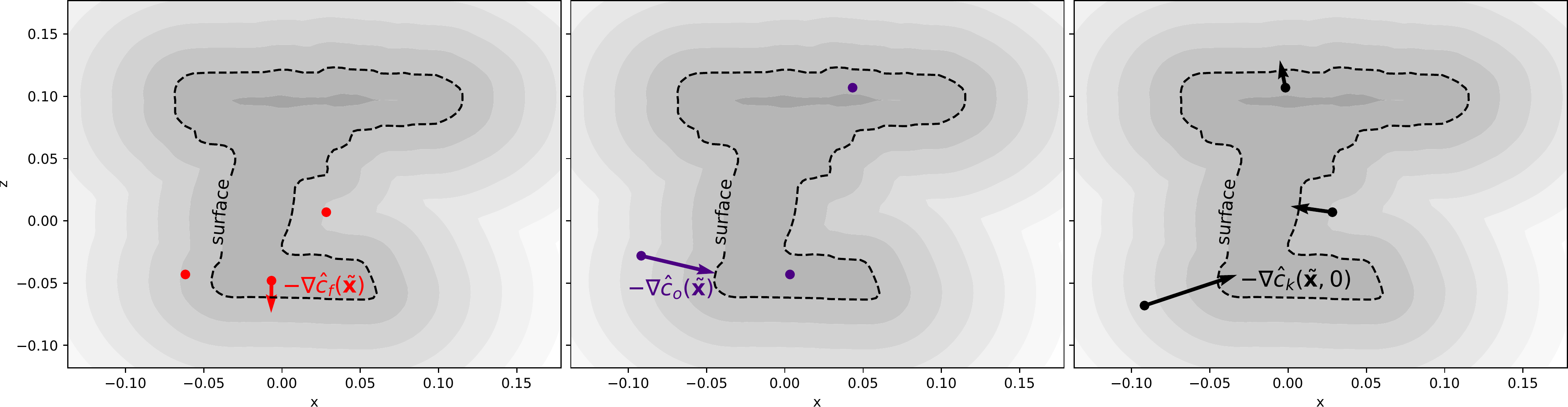}
  \caption{$\relaxedtotalcost(\xall, \T)$ negative gradients with respect to sampled points shown for a X-Z cross section of the YCB drill. Red points are known free space and $-\nabla \freecost$  pushes them outside the object. Purple points are known occupied and $-\nabla \occupiedcost$ pushes them inside the object. Black points have known SDF values (here they are known surface points, $\sdf(\xtrans) = 0$) and $-\nabla \knowncost$ pushes them towards the corresponding SDF level set.} 
  \label{fig:costs}
\end{figure*}

\section{Method}
This section presents \methodabv{}, which consists of a differentiable cost function (relaxation of Eq.~\ref{eq:gtcost}) that enables gradient-based methods to reduce a transform's cost, and a quality diversity optimization scheme, which uses that cost to estimate the set of plausible transforms. We also show how to update \methodabv{}'s $\estimateset{}$ estimates online as more points are perceived.

\subsection{Relaxation of Semantic Constraints}
Eq.~\ref{eq:gtcost} has discrete components and is not differentiable. We would like a relaxation $\relaxedtotalcost(\xall, \T)$ of $\totalcost(\xall, \T)$ that is differentiable to be more amenable to optimization. For convenience, when the $\T$ used is unambiguous, we denote point positions in the world frame transformed to an estimated object frame as $\xtrans$, where 
$\xtrans = \T\x$ in homogeneous coordinates (append 1 to $\xtrans$ and $\x$).
Specifically, we want to efficiently compute the gradient $\nabla_\T \relaxedtotalcost(\xall, \T)$. For better geometric intuition, we consider the gradient contributed by each known point:
\begin{equation}
\nabla_\T \relaxedtotalcost(\xall, \T) = \sum_{n=1}^\numpoints \nabla_{\T} \relaxedcost(\T\x_\n, \s_\n) = \sum_{n=1}^\numpoints \nabla_{\xtrans} \relaxedcost(\xtrans, \s_\n)
\end{equation}
This gradient is spatial and with respect to the transformed point. Intuitively, gradient descent will move the points spatially along their negative gradients through adjusting $\T$. This is visualized in Fig.~\ref{fig:costs}. As it is clear what each gradient is with respect to, we drop the subscript in future usage.

The separate semantic classes motivate us to consider each case separately. We partition $\xall$ into $\xfree = \{(\x,\s) \ | \ \s = \free \}$, $\xocc = \{(\x,\s) \ | \ \s = \occupied\}$, and $\xknown = \{(\x,\s) \ | \ \s \in \reals\}$. We then decompose the gradient: 
\begin{align}\label{eq:cost}
    \nabla \relaxedtotalcost(\xall, \T) = \sum_{\mathclap{\x,\s \in \xfree}} \nabla \freecost(\xtrans) + \sum_{\mathclap{\x,\s \in \xocc}} \nabla \occupiedcost(\xtrans) + \sum_{\mathclap{\x,\s \in \xknown}} \nabla \knowncost(\xtrans, \s)
\end{align}
At each point, the cost arises from an SDF value mismatch and thus the gradient must be along the direction of greatest SDF value change. This is provided precisely by $\nabla_\xtrans \sdf(\xtrans)$, the gradient (normalized such that $\norm{\nabla \sdf(\xtrans)}_2 = 1$) of the SDF at that point. Thus all cost gradients must be parallel or anti-parallel to the SDF gradient. See Section~\ref{sec:sdfquery} for how we achieve efficient lookup of SDF values and gradients. Fig.~\ref{fig:costs} shows our cost applied to points of each semantic class. Arrows indicate the negative cost gradient experienced by that point, which is the spatial direction the points will move along when we perform gradient descent on the cost. We define the gradients directly and assign its magnitude as the cost value.

\subsubsection{Free space cost}
From Eq.~\ref{eq:gtcost}, points in $\xfree$ achieve 0 cost when $\sdf(\xtrans) > 0$. When $\sdf(\xtrans) \leq 0$ the negative gradient points towards the SDF 0-level set (surface of the object). To tolerate small degrees of violation due to uncertainty in the point positions, we aim for the $\interiorthreshold$-level set where $\interiorthreshold < 0$. We define the magnitude of free space violation as $\max(0, \interiorthreshold-\sdf(\xtrans))$. This has the effect of only giving non-zero gradients to violations beyond $\interiorthreshold$. Thus we have
\begin{align}
       \nabla \freecost(\xtrans) = - \scalefree \max(0, \interiorthreshold-\sdf(\xtrans)) \nabla \sdf(\xtrans)
\end{align}
where $\scalefree > 0$ is a scaling parameter. In a sense, it controls the degree of relaxation since using smaller values will lead to a smoother optimization path,
particularly near the start of the optimization, while a higher value is needed to enforce the high cost from Eq.~\ref{eq:gtcost}. This scaling parameter can be annealed during the optimization process, i.e. starting with a small value and increasing over optimization iterations.

\subsubsection{Occupied space cost}
Symmetric to the free space cost, violating occupied points moves along $-\nabla \occupiedcost$ to the $-\interiorthreshold$-level set. In this case, violation occurs when $\sdf(\xtrans) > -\interiorthreshold$ and has magnitude $-\min(0, -\interiorthreshold -\sdf(\xtrans))$: 
\begin{align}
       \nabla \occupiedcost(\xtrans) = -\scalefree \min(0, -\interiorthreshold -\sdf(\xtrans)) \nabla \sdf(\xtrans)\\
       = \scalefree \max(0, \interiorthreshold + \sdf(\xtrans)) \nabla \sdf(\xtrans)\nonumber
\end{align}

\subsubsection{Known SDF cost}
This cost is a generalization of surface matching present in many registration methods. Known surface points are a special case of $\s = 0$, and is commonly perceived through contact and visual perception.
The cost's structure is similar to the previous costs, with the difference being that instead of $\interiorthreshold$ and $-\interiorthreshold$, each point has a separate desired level set given by its semantic value:
\begin{align}
       \nabla \knowncost(\xtrans, \s) = (\sdf(\xtrans) - \s) \nabla \sdf(\xtrans)
\end{align}

\subsection{SDF Query Improvements}\label{sec:sdfquery}
Evaluating $\nabla \relaxedcost(\xall, \T)$ requires computing $\sdf(\xtrans)$ and $\nabla \sdf(\xtrans)$ for $\numpoints$ known points.
Regardless of the structure and efficiency of the given $\sdf$, we precompute a voxel-grid approximation of it to enable fast parallel lookup. Each voxel reports the SDF value and gradient at the center of it. Each voxel is cubic with side length (resolution) $\rtarget$, with the whole grid being the object's bounding box padded by $\sdfinflation > 0$ on all sides. Queries of points outside the voxel-grid are deferred to the original $\sdf$, with $\sdf(\xtrans) = ||\xtrans - \x'||_2$ and $\nabla_\xtrans (\xtrans) = (1 - 2\indicator(\text{$\xtrans$ inside $\mesh$})) (\xtrans - \x') $. Where $\x' = \argmin_{\x \in \mesh} ||\xtrans - \x||_2$ is the closest point on the mesh to $\xtrans$, and a ray is traced from the inside of the object (assuming object-centered origin is interior) to $\xtrans$, with an even number of mesh surface crossings indicating it is inside. Lower $\rtarget$ (a denser grid) trades higher memory usage for more accurate representation.

Another challenge to the efficiency of evaluating $\nabla \relaxedtotalcost(\xall, \T)$ is the representation of known free points $\xfree$. This is typically a volume, such as the space swept out by a robot's motion or derived from visual data. Representing this volume as a dense set of points makes $\nabla \freecost$ prohibitively expensive to evaluate. Similar to the 3D Normal Distribution Transform~\cite{magnusson2009three}, we discretize the free space into a voxel-grid. The voxel-grid has resolution $\rfree$, and the whole grid expands to the range of free points. $\rfree$ allows us to set the maximum point density.

\subsection{Quality Diversity Optimization}
With Eq.~\ref{eq:cost} we can optimize an initial $\T$ using stochastic gradient descent (SGD). The optimized $\T$ depends on the starting $\T$ and will achieve a local minima of $\relaxedtotalcost$. A naive approach to creating the estimated plausible transform set $\estimateset$ is to start with a set of transforms $\estimateset_0$ and perform SGD on each $\T \in \estimateset_0$ separately. 
We compare to this approach as an ablation in our experiments, where we find that this method often produces $\estimateset$ with poor plausible diversity as it relies only on the diversity of local minima for coverage. 

Instead, we turn to Quality Diversity (QD) optimization. At a high level, in addition to the $\reals^m$ solution space to search over to maximize an objective $f: \reals^m \rightarrow \reals$, there are $k$ behavior (also known as measure) functions $\imeasure_i : \reals^m \rightarrow \reals$,  jointly $\measure : \reals^m \rightarrow \reals^k$. For the behavior space $\measurespace = \measure(\reals^m)$ (image of $\measure$), the QD objective is to find for each $\measurevalue \in \measurespace$ a solution $\solution \in \reals^m$ such that $\measure(\solution) = \measurevalue$ and $f(\solution)$ is maximized. See~\citet{pugh2016quality} for an overview of the field.

For our problem, $f(\solution) = -\relaxedtotalcost(\xall, \T)$, and we search over the transforms represented in $\reals^9$, with 3 translational components and the 6 dimensional representation of rotation suggested by~\citet{zhou2019continuity}.
Our $\measure$ extracts the translational components of the pose. In a sense, we are searching for the best rotation given some translation to minimize $\relaxedtotalcost$. 
Intuitively, QD's enforced diversity over $\measurespace$ will prevent the collapse of $\estimateset$ when $\xall$ does not sufficiently constrain our estimation.

In particular, we use CMA-MEGA~\cite{fontaine2021differentiable} optimization to take advantage of our cost's differentiability to more efficiently search for good solutions. $\measurespace$ is discretized into a regularly spaced grid, called the archive, with each cell holding the best solution for that cell. Diversity in $\estimateset$ is enforced by requiring each $\T \in \estimateset$ to come from a different cell in $\measurespace$. This is an evolutionary method in that the lowest cost transforms from different cells are iteratively combined to generate new transforms. Thus, it is valuable to populate the archive with low cost transforms $\estimateset_l$ to initialize the search. If no prior estimate is available, $\estimateset_l = \{\}$, but when we run \methodabv{} iteratively, $\estimateset_l$ contains the estimates from the previous iteration (see Sec. \ref{sec:online}).

\newcommand{\bm}{\boldsymbol{\mu}}
\newcommand{\bs}{\boldsymbol{\sigma}}
Algorithm~\ref{alg:qd} describes how we use QD optimization. First, we run SGD on the given initial transform set $\estimateset_0$ using Eq.~\ref{eq:cost} to create an $\estimateset'$. We extract its translation components $\measure(\estimateset) = \positionset$. Using the mean $\bm$ and standard deviation $\bs$ of $\positionset$ along each dimension, we size the grid $\measurespace$ between $[\bm - \binsigma\bs, \bm + \binsigma\bs]$. The grid is centered on the mean $\bm$ with extents scaled by the standard deviation $\bs$ along each dimension.
A large $\bs$ suggests that there are low cost solutions with very different values along that dimension, motivating a wider search range. The parameter $\binsigma > 0$ adjusts how many standard deviations out we search for solutions. 
We initialize $\measurespace$ with known low cost transforms from $\estimateset_l$, along with the SGD solutions $\estimateset'$ to seed the QD optimization.
Note that sizing the archive defines the region of the behavior space to search over while initializing it populates some grid cells with transform values.
We then run CMA-MEGA on $\measurespace$ for $\qditernum$ iterations to populate $\measurespace$ with the lowest cost $\T$ for each cell. Finally, we select the $\T$ from the $|\estimateset_0|$ lowest cost cells as $\estimateset$.

Since we initialize $\measurespace$ with $\estimateset' \cup \estimateset_l$, the
QD optimization can be seen as a fine-tuning process. Initially, each $\T \in \estimateset' \cup \estimateset_l$ is the best solution for their respective cells (assuming they fall into different cells of $\measurespace$). If a lower cost transform exists for a cell, QD optimization will eventually find it and replace the original $\T$ from $\estimateset' \cup \estimateset_l$.

\begin{algorithm}[t]
\DontPrintSemicolon
\SetKwInput{Given}{Given}
\SetKwInput{Hyperparameters}{Hyperparameters}

\Given{
$\xall$ known points,
$\estimateset_0$ initial transform set,
$\estimateset_l$ low cost transform set,
$\binsigma$ number of standard deviations to consider, $\qditernum$ number of QD iterations
}
$\estimateset' \leftarrow $ SGD on $\estimateset_0$ with $\relaxedtotalcost(\xall, \T)$\;
$\positionset \leftarrow \measure(\estimateset')$\;
$\bm \leftarrow $ mean($\positionset$), $\bs \leftarrow$ std($\positionset$)\;
$\measurespace \leftarrow $ grid with dimensions $[\bm - \binsigma\bs, \bm + \binsigma\bs]$\;
$\measurespace \leftarrow \text{UpdateCells}(\measurespace, \estimateset' \cup \estimateset_l)$ \label{line:addsolution} \tcp{initialize the search with low cost transforms}
$\measurespace \leftarrow $
 CMA-MEGA$(\measurespace, \relaxedtotalcost, \qditernum)$\;
$\estimateset \leftarrow \{\T | \T \in $ cells from $\measurespace$ with $|\estimateset_0|$ lowest costs $\}$\;
\caption{\methodabv{}: QD optimization for $\estimateset$}
\label{alg:qd}
\end{algorithm}

\subsection{Online Updates to $\estimateset$}\label{sec:online}
Registration can be performed iteratively as new sensor data are added to $\xall$. Information from the previous registration allows us to more efficiently search for $\estimateset$. Our update process is described in Algorithm~\ref{alg:update}.
First, we consider the generation and update of the initial transform set $\estimateset_0$. Before any registration, we sample uniformly at random (both position and rotation), within the given workspace  $\workspace$ where the object could possibly be. Assuming the object has not moved, we update $\estimateset_0$ as $|\estimateset_0|$ perturbations around the best $\T$ of the previous estimated set, $\T' = \argmin_{\T \in \estimateset}\relaxedtotalcost(\xall,\T)$. We perturb its translational components with Gaussian noise $\updatenoiset > 0$ along each dimension. Then, in line~\ref{line:axissample}, we uniformly sample a rotation axis, scaled with an angle sampled as Gaussian noise with $\updatenoiser > 0$. We take the exponential map of this axis-angle representation and multiply it by the rotational component of $\T'$. This sampling based update of $\estimateset_0$ helps with escaping bad local minima.

Secondly, the data update changes $\xall$ and invalidates the previously computed $\measurespace$, but the solutions in each cell of the $\measurespace$ may still have low cost with respect to the updated $\xall$. We select them as $\estimateset_l$ and use them to initialize the next QD optimization in line~\ref{line:addsolution} of Algorithm~\ref{alg:qd}.

\newcommand{\dt}{\boldsymbol{\Delta t}}
\newcommand{\dr}{\boldsymbol{\Delta R}}
\begin{algorithm}[t]
\DontPrintSemicolon
\SetKwInput{Given}{Given}
\SetKwInput{Hyperparameters}{Hyperparameters}

\Given{
$\workspace$ workspace dimension,
$\updatenoiset$ translation noise,
$\updatenoiser$ rotation noise
}
$\estimateset_0 \leftarrow  \positionset \sim U(\workspace) \times U(\rotset)$\;
$\estimateset_l \leftarrow \{\}$\;
\While{register}{
$\xall \leftarrow $ perceived from environment\;
$\estimateset \leftarrow $ \methodabv{}$(\xall, \estimateset_0, \estimateset_l)$\;
$\estimateset_l \leftarrow \estimateset$\;
$\T' \leftarrow \argmin_{\T \in \estimateset}\relaxedtotalcost(\xall, \T)$\;
$\estimateset_0' \leftarrow \{\}$\;
\For{$i \leftarrow 1$ \KwTo $|\estimateset_0|$}{
    $\dt \sim \normal(\zero, \diag([\updatenoiset, \updatenoiset, \updatenoiset])$\;
    $\theta \sim \normal(0, \updatenoiser)$\;
    $\mathbf{e} \sim U(\{\x \ | \ \norm{\x}_2 = 1, \x \in \reals^3\})$\label{line:axissample}\;
    $\dr \leftarrow e^{\theta \mathbf{e}}$\tcp{axis angle to matrix}
    $\estimateset_0' \leftarrow \estimateset_0' \cup \{ (\dt + \text{trans}(\T'), \dr  \cdot \text{rot}(\T'))\}$
}
$\estimateset_0 \leftarrow \estimateset_0'$\;
}
\caption{Online update of $\estimateset$}
\label{alg:update}
\end{algorithm}
\section{Experiments}
\label{sec:results}
In this section, we first describe our simulated and real robot environments, and how we estimate $\xall$ from sensor data. Next, we describe how we generate the plausible set. Then, we describe our baselines and ablations and how we quantitatively evaluate each method on probing experiments of objects in simulation and in the real world. Lastly, we evaluate the value of our cost as a QD objective.

In these experiments, a target object is in a fixed pose inside an occluded cabinet, and we estimate its pose through a fixed sequence of probing actions by a robot (some of which will result in contact). 
We run all methods after each probe, updating our set of pose estimates using Algorithm~\ref{alg:update}. Note that all methods receive the same known points $\xall$ after each probe. Each probe extends the robot straight into the cabinet for a fixed distance or until contact. The probing locations are configuration specific and designed such that at least some contacts are made. 
We use YCB objects~\cite{calli2017yale} for both simulated and real experiments as their meshes are readily available. We precompute the object-frame SDF from these meshes as described in Section~\ref{sec:sdfquery}. In all cases, we are estimating a set of 30 transforms ($|\estimateset|=30$), and use parameters $\interiorthreshold=-10mm, \binsigma=3, \scalefree=20, \updatenoiset=0.05m, \updatenoiser=0.3$. In the sim experiments we use $\qditernum=100$ and in the real world $\qditernum=500$. 
For the Plausible Diversity distance function $\tdist(\T, \estT)$, we sample 200 points on the object surface and evaluate the Chamfer Distance between them after being transformed. Note that the 200 points sampled are different per trial. We extract the $x$ and $y$ components of the pose using $\measure$ - we found no significant difference in performance from also extracting $z$.

\begin{figure}
    \centering
    \includegraphics[width=0.49\linewidth]{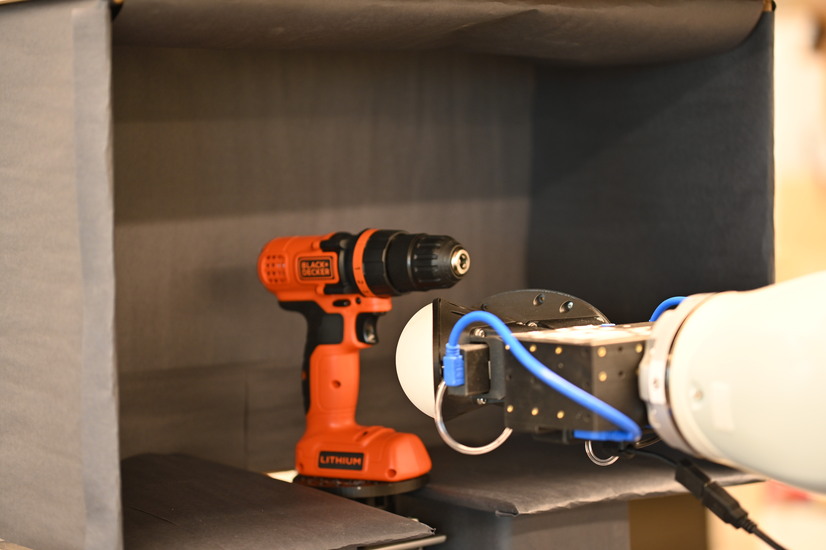}
    \includegraphics[width=0.49\linewidth]{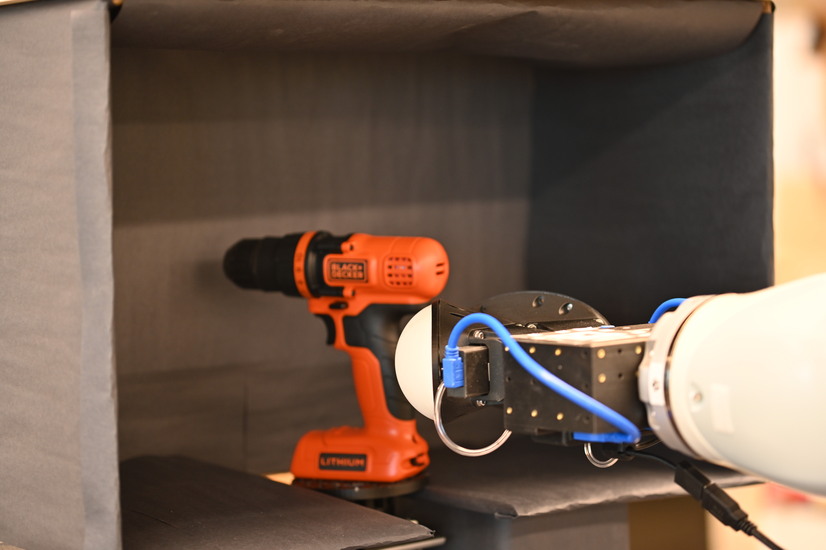}
    \includegraphics[width=0.49\linewidth]{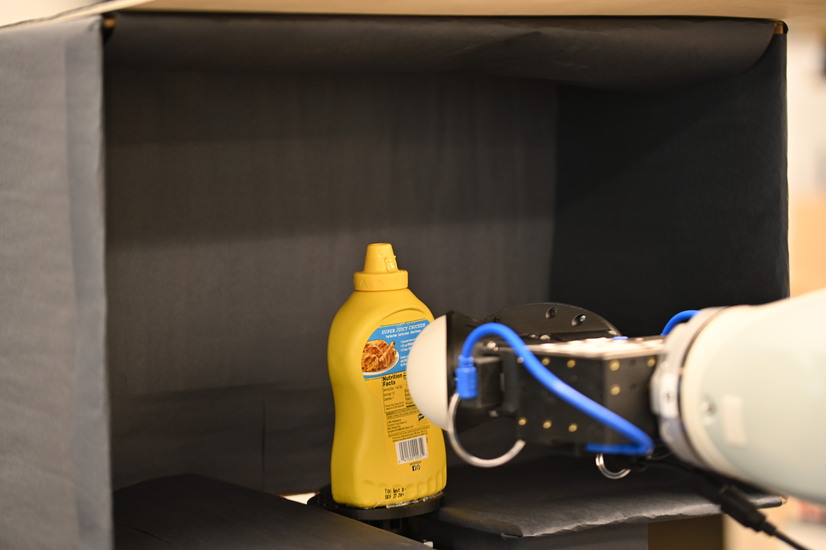}
    \includegraphics[width=0.49\linewidth]{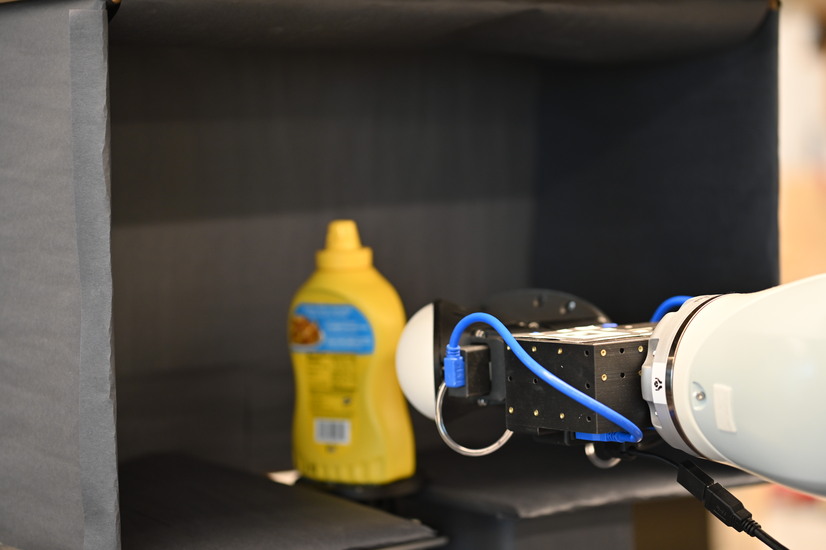}
    \caption{Real probing configurations for the drill and mustard.}
    \label{fig:realconfig}
\end{figure}

\begin{figure}
    \centering
    \includegraphics[width=\linewidth]{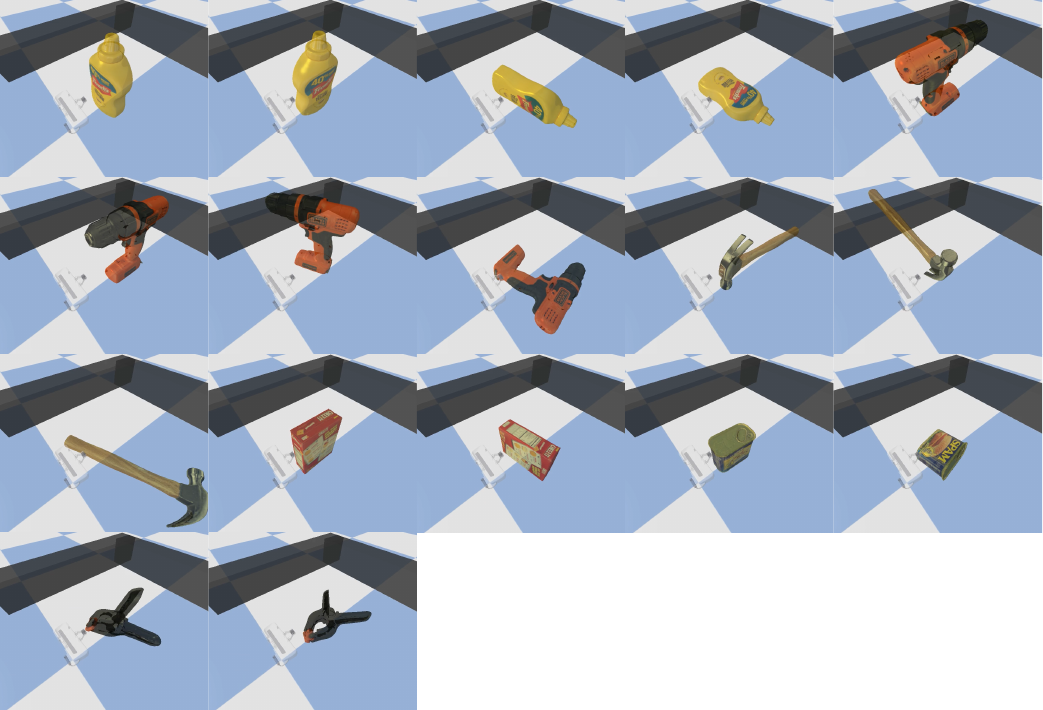}
    \caption{Simulated probing configurations for multiple YCB objects.}
    \label{fig:simconfig}
\end{figure}

\begin{figure}[h]
\includegraphics[width=0.45\linewidth]{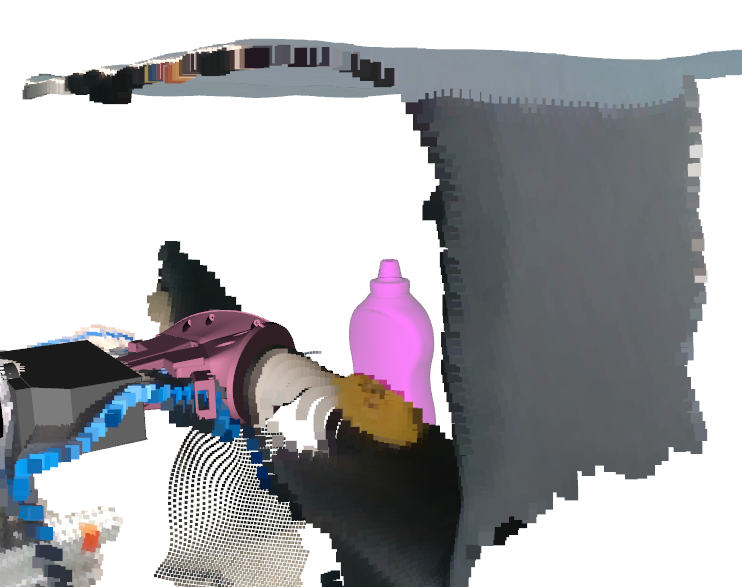}
\includegraphics[width=0.45\linewidth]{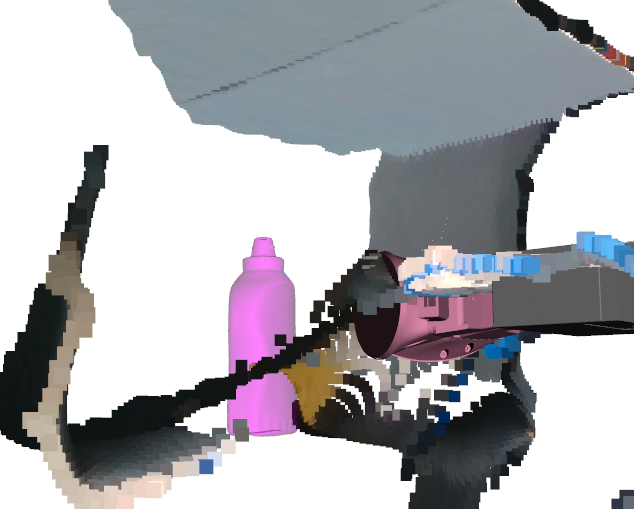}
   \caption{Unreliable RGBD readings inside the partially occluded cabinet, viewed from both sides, with an approximate pose of the mustard bottle in purple.}
  \label{fig:unreliable}
\end{figure}
\begin{figure*}
    \centering
    \includegraphics[width=\linewidth]{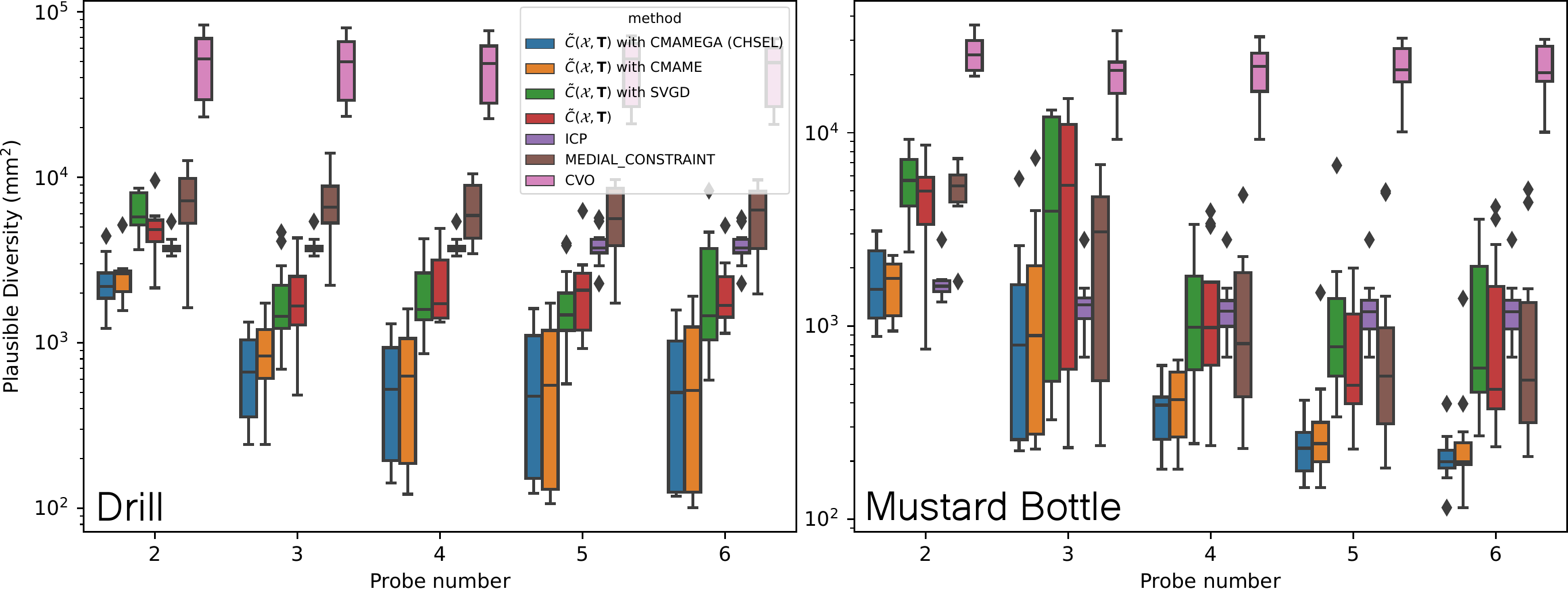}
    \caption{Plausible Diversity for real drill (left) and mustard (right) probing experiments across 2 configurations and 6 trials each. The bars indicate the 25 to 75 percentile while the whiskers are the min and max with outliers as diamonds. Lower is better.}
    \label{fig:realres}
\end{figure*}

\begin{figure*}
    \centering
    \includegraphics[width=\linewidth]{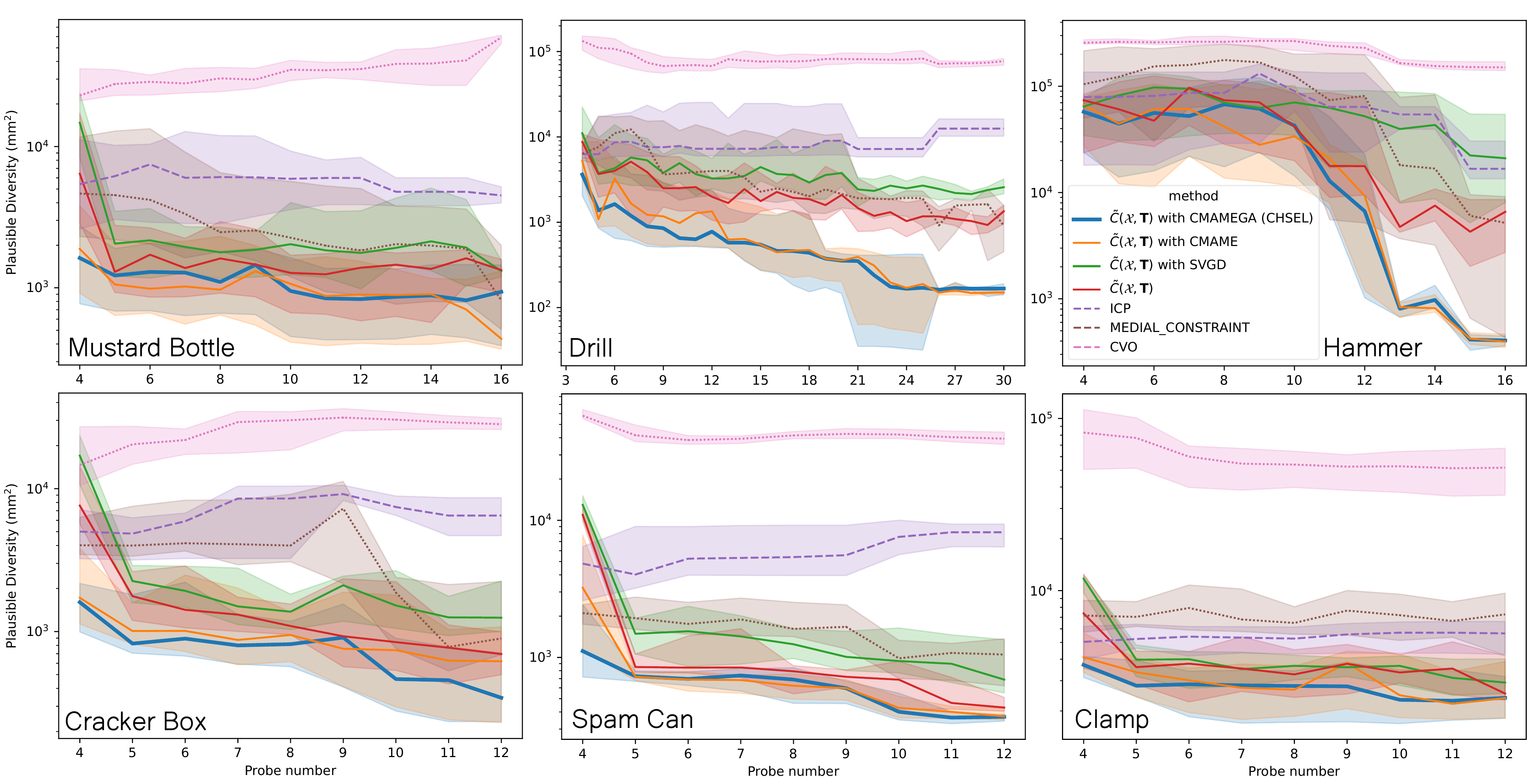}
    \caption{Plausible Diversity for simulated probing experiments across different YCB objects, with 2 to 4 configurations from Fig.~\ref{fig:simconfig} over 10 trials each. The median is in bold while the shaded region represents 25 to 75 percentile. Lower is better.}
    \label{fig:simres}
\end{figure*}

\subsection{Simulated Environment}
We use PyBullet~\cite{coumans2016pybullet} to simulate a Franka Emika (FE) gripper (see Fig.~\ref{fig:simconfig}) that is position controlled. 
The workspace is voxelized with resolution $\rfree=25mm$ and spans $[-0.1, 0.5] \times [-0.3, 0.3] \times [-0.075, 0.625]$ in meters. We label the boundary of the workspace as free space. 
The SDF is voxelized with resolution $\rtarget=10mm$ with padding $\sdfinflation=50mm$. The robot sweeps out voxels in the workspace grid during its probing motions, and $\xfree$ is given by the center of the swept voxels. $\xocc$ is empty as we have have no sensors that detect non-surface occupancy, though such information can be added if available. $\xknown$ is given by the contact points, with each having semantics $\s = 0$ since contact can only occur on the object surface. Both the gripper and object are rigid and so only make single-point contacts which we retrieve from the simulator. For different trials, we seed the random number generator with different values.

\subsection{Real Environment}

For our real world experiment, we equip a 7DoF KUKA LBR iiwa arm with two soft-bubble tactile sensors~\cite{kuppuswamy2020soft} on the gripper (see Fig.~\ref{fig:teaser} and Fig.~\ref{fig:realconfig}).
The soft-bubble sensors allow us to detect patch contact, which we consider as any point with deformation beyond $4mm$ and being in the top $10^{th}$ percentile of deformations. We use a mean filter to remove noise and downsample such that each contact produces at most 50 surface points.

The workspace is a physical cabinet mock-up and is voxelized with resolution $\rfree=10mm$, spanning $[0.7, 1.1] \times [-0.2, 0.2] \times [0.31, 0.6]$ meters. The SDF is voxelized with resolution $\rtarget=5mm$ with padding $\sdfinflation=50mm$. In addition to populating $\xfree$ with robot swept volume, we utilize a RealSense RGBD camera, partially occluded by the cabinet. The camera is unreliable near occluding edges (see Fig.~\ref{fig:unreliable}), thus we do not assume the object can be reliably segmented from the camera view, so we only use the free space information derived from the depth data. To that effect, we trace rays from the camera to 95\% of each pixel's detected depth and add them to $\xfree$.

\subsection{Computing Plausible Set}
In order to evaluate our method, we need to compute $\plausibleset_\suboptimality$, which is very computationally intensive. We compute $\plausibleset_\suboptimality$ by densely sampling transforms around $\T^*$ and evaluate each using Eq.~\ref{eq:gtcost} with $\cmax=100000$. Specifically, we search over a grid spanning $[-0.1,0.15]\times[-0.2,0.2]\times[0,0.1]$ meters with 15 cells along each dimension. We also uniformly random sample 10000 rotations which we combine with each translation cell. See Table~\ref{tab:suboptimality} for the $\suboptimality$ used to generate the plausible set of each object. They were selected such that most probe trials have around 30 members in $\plausibleset_\suboptimality$ halfway through.

In simulation, we retrieve $\T^*$ from the simulator, while on the real robot we first manually specify an approximate pose, then search in two passes. The first pass searches around the specified pose to find the optimal transform, which is then used as $\T^*$ for the second pass.

\begin{table}[]
    \centering
\begin{tabular}{|l|l|}
\hline
Object              & $\suboptimality$ \\ \hline
Real Drill          & 0.001          \\
Real Mustard Bottle & 0.0003         \\
Sim Drill           & 0.001          \\
Sim Mustard Bottle  & 0.0003         \\
Sim Hammer          & 0.001          \\
Sim Cracker Box     & 0.0005         \\
Sim Spam Can        & 0.0003         \\
Sim Clamp           & 0.0007         \\ \hline
\end{tabular}
    \caption{$\suboptimality$ used to generate the plausible set for each objects.}
    \label{tab:suboptimality}
\end{table}

\subsection{Baselines}
We compare against \textbf{ICP} as a weak baseline that does not use free space information. ICP registers the known surface points against another point set, which we provide as 500 points randomly sampled from the object surface. Note that these points are different for each trial. ICP is run until convergence.

Secondly, we compare against Continuous Visual Odometry (\textbf{CVO})~\cite{zhang2021new}, the state of the art in semantic point set registration, and a continuous generalization of 3D NDT. We use 2 dimensional semantics to represent free points as $[0.9, 0.1]$ and surface points as $[0.1, 0.9]$. CVO registers the free and surface points against another semantic point set, which we provide as the center of the precomputed SDF voxels. Voxels with SDF value between $[-\rtarget, \rtarget]$ are labelled with surface semantics, and voxels with SDF value greater than $\rtarget$ are labelled as free. Note that there are many more free points than surface points ($\approx 125:1$). 

Next, we consider using Stein Variational Gradient Descent (\textbf{SVGD})~\cite{liu2016stein} of Eq.~\ref{eq:cost} to enforce diversity. We formulate $p(\T|\xall) \propto e^{-\beta \relaxedtotalcost(\xall, \T)}$, and select $\beta=5$ as a scaling term for how peaked the distribution is. We have $|\estimateset_0|$ stein particles, each one initalized with a separate $\T \in \estimateset_0$, implicitly defining the prior. We use an RBF kernel with scale 0.01. 

\newcommand{\bc}{B_c}
Lastly, we compare against~\citet{haugo2020iterative}, which forms free space constraints by covering the free space using balls along the volume's medial axis (we refer to this baseline as \textbf{Medial Constraint}). For each ball we have cost $\max(0, B_r - \sdf(\Tilde{\bc}))^2$ where $B_r$ is its radius and $\bc$ is the center position of the ball. For each surface point we have cost $\sdf(\xtrans)^2$. The total cost is the sum of the mean ball cost and the mean point costs. We optimize this cost using CMA-ES~\cite{hansen2003reducing}, a gradient-free evolutionary optimization technique.

\subsection{Ablations}
We ablate components of our method starting with how useful the gradient is for accelerating QD optimization. Instead of CMA-MEGA, we use \textbf{CMA-ME}~\cite{fontaine2020covariance} which does not explore using gradients.

We also consider just gradient descent on Eq.~\ref{eq:cost} to evaluate the value of additional optimization. We run Adam for 500 iterations with learning rate 0.01, reset to 0.01 every 50 iterations. These are also the parameters used for initializing CMA-ME and CMA-MEGA.

\begin{figure}
    \centering
    \includegraphics[width=\linewidth]{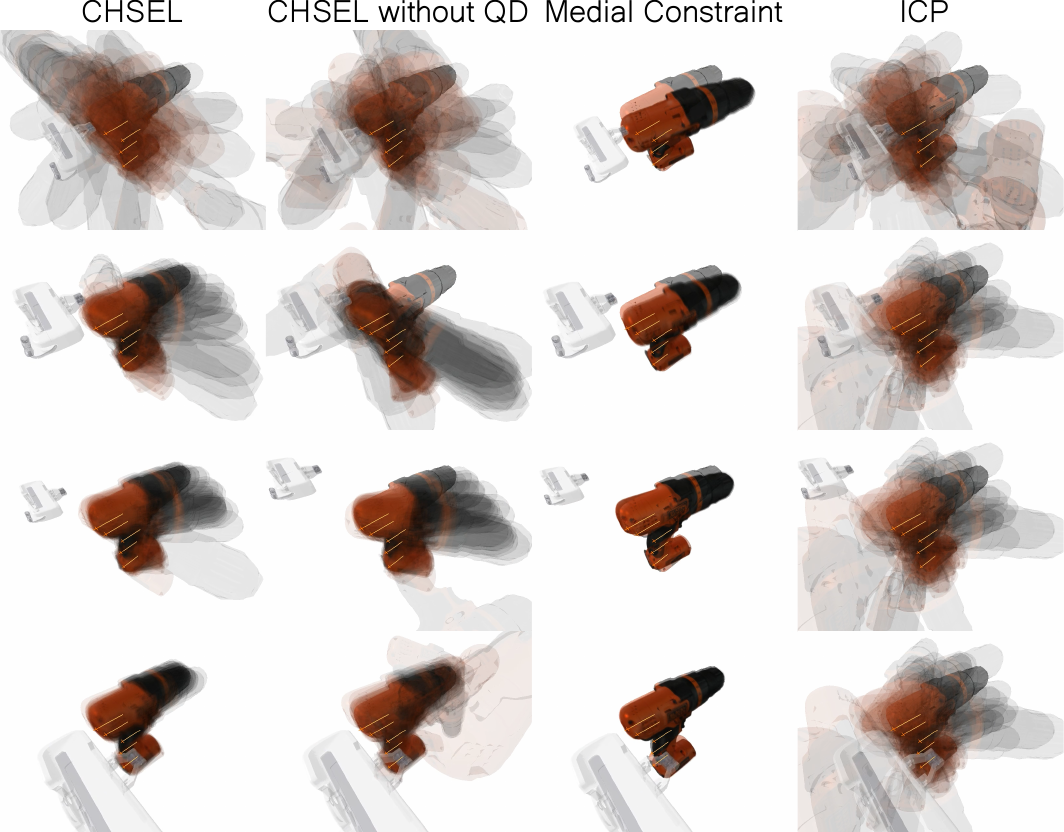}
    \caption{Reducing uncertainty in estimated pose as a result of additional free space points for selected methods, obtained by probing to the sides of the YCB drill. $\estimateset$ is represented as transformed copies of the mesh while contact points are drawn in orange, with the line indicating the direction of the probe.}
    \label{fig:example}
\end{figure}
\subsection{Probing Experiments}
Qualitatively, we see the progress of a probing experiment and the elimination of hypothesis transforms through gaining known free space points in Fig.~\ref{fig:example}. From the initial probes along the back of the drill, it could take on many possible upright orientations. Note that after the probe in the second row, the contact points constrain the pose such that the contacts must lie on the back of the drill. As we probe the left side of the drill, without making contact, we eliminate transforms that would conflict with the new free space points. Probing the other side further narrows down the plausible transforms. Note the lack of diversity from the Medial Constraint baseline and the poor estimation from ICP since it cannot use free space information.

\begin{figure}
    \centering
    \includegraphics[width=\linewidth]{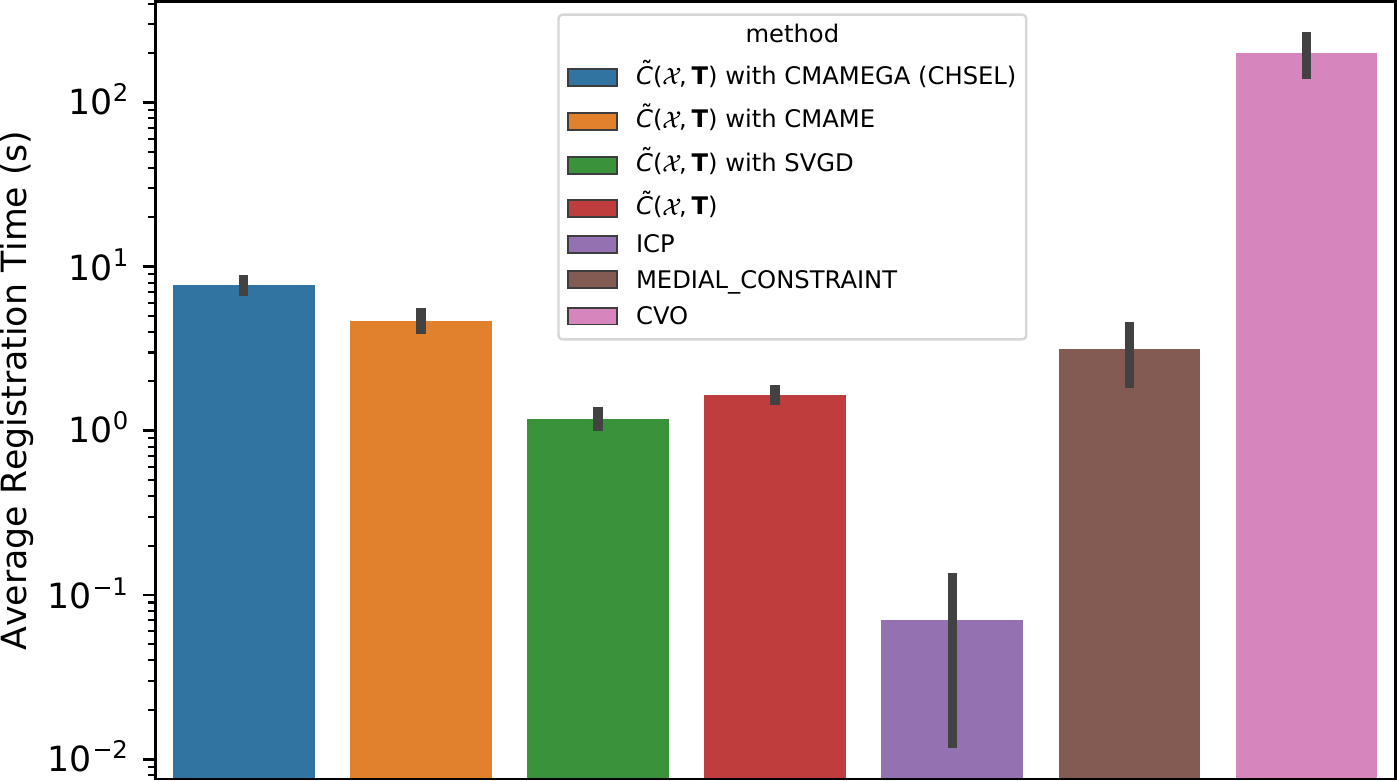}
    \caption{Average time per registration of 30 transforms on the real probing experiments. Error bars indicate one standard deviation.}
    \label{fig:realtime}
\end{figure}

Fig.~\ref{fig:realres} summarizes the results of the real probing experiments on the YCB drill and mustard bottle, each in two different configurations (see Fig.~\ref{fig:realconfig}) over 6 trials.  Fig.~\ref{fig:simres} summarizes the results of the simulation probing experiments on the YCB drill, mustard bottle, hammer, cracker box, spam can, and clamp. Additionally, we show the average time it takes for each method to perform registration on the real experiments in Fig.~\ref{fig:realtime}. This involves producing 30 transforms with $|\xall| \ \in [13000, 21000]$, and $|\xknown| \ \in [0, 150]$. Note that all methods apart from CVO use parallelized implementations.
Computations were performed on a NVIDIA GeForce RTX 2080 Ti with 11GB of VRAM.

From Fig.~\ref{fig:realres} and Fig.~\ref{fig:simres}, we see that applying QD optimization to $\relaxedtotalcost$ in general outperforms baselines and the ablations. This is particularly true on more irregular objects such as the drill and hammer, and when we have noisy data in the real experiment. Even without QD optimization, gradient descent on $\relaxedtotalcost$ outperforms the Medial Constraint baseline. This may be due to the ability of CMA-ES to escape local minima, leading to low coverage, as seen in Fig.~\ref{fig:example}. All methods, including ICP, outperform CVO. We suspect this is due to the large imbalance of $|\xfree|$ to $|\xknown|$ ($\approx 125:1$). See Appendix~\ref{sec:cvoperf} for an investigation of CVO's performance.

\subsection{QD Method Comparison}
\begin{figure}
    \centering
    \includegraphics[width=\linewidth]{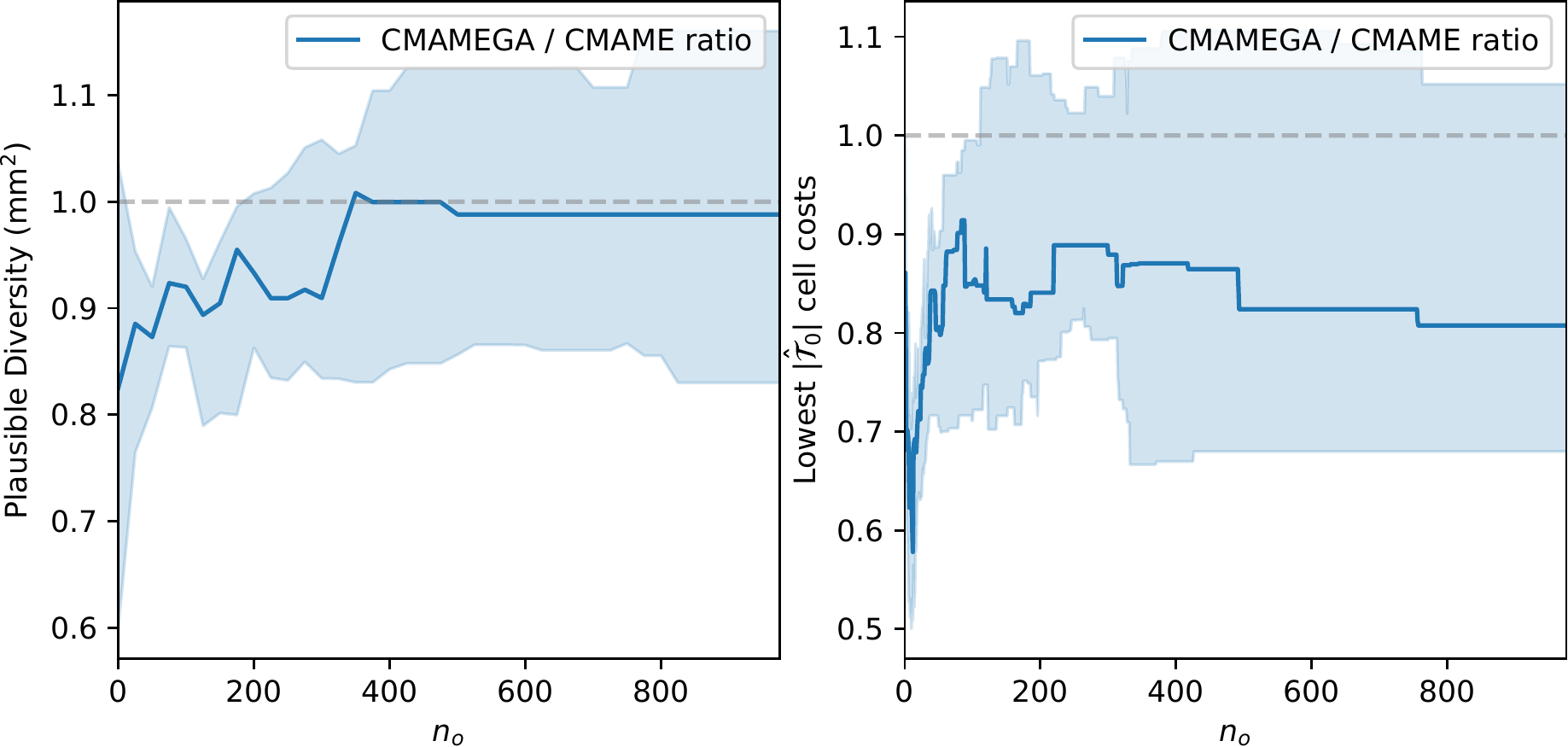}
    \caption{Comparison of QD optimization progress using $\relaxedtotalcost$ on the real drill experiment. Results are averaged across 6 trials and probe numbers 5 and 6. Median is in bold while the shaded region represents 25 to 75 percentile.}
    \label{fig:qdcompare}
\end{figure}

We investigate the value of our formulated cost's differentiability by considering the QD optimization process in further detail. In Fig.~\ref{fig:qdcompare}, we compare the $\estimateset$ performance of using CMA-MEGA and CMA-ME as we increase the number of QD optimization iterations. The results are from the front-facing real drill experiment (Fig.~\ref{fig:realconfig} top left), averaged over probes 5 and 6, and across the 6 trials. Both methods are initialized with the same $\estimateset_0$ and $\estimateset_l$ each trial and probe (see Algorithm~\ref{alg:qd}). We see that CMA-MEGA is able to use our gradients to reach lower Plausible Diversity and average cost of the best cells in fewer iterations, and that they converge and reach parity after around 500 iterations (fewer in simulation due to lack of noise). In Fig.~\ref{fig:realres} and Fig.~\ref{fig:simres}, both methods have run for enough iterations to converge. On average, each CMA-ME iteration takes 8.37ms while each CMA-MEGA iteration takes 11.5ms.

\subsection{QD Objective Comparison}
\begin{figure}
    \centering
    \includegraphics[width=\linewidth]{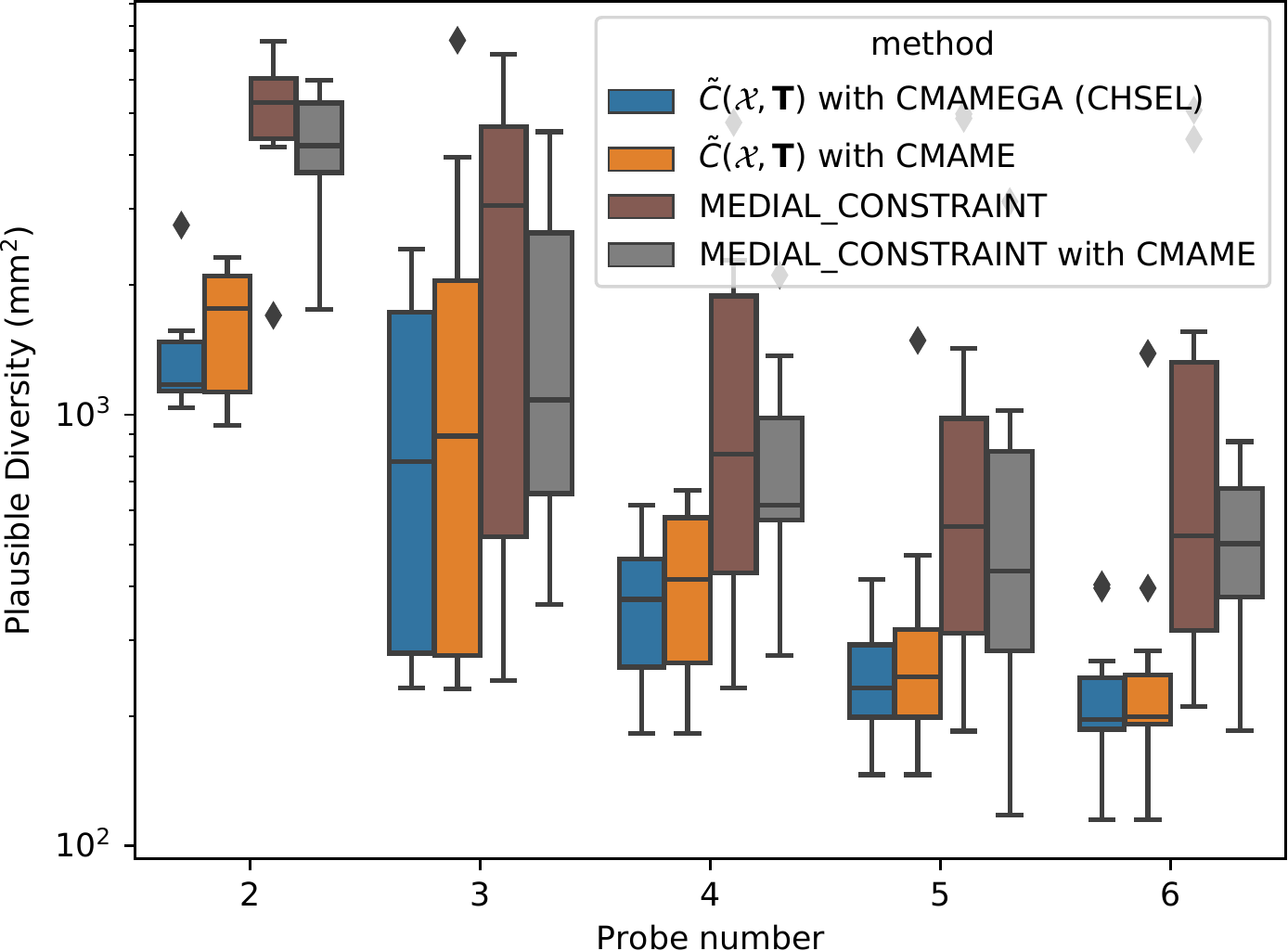}
    \caption{Plausible Diversity on the two real mustard bottle experiments with a focus on the improvement the Medial Constraint baseline receives from QD optimization.}
    \label{fig:qdobjective}
\end{figure}

Lastly, we investigate how well QD optimization works with other objectives. We perform CMA-ME optimization using the Medial Constraint objective, with $\measurespace$ initialized and sized from the $\estimateset$ estimated by Medial Constraint using CMA-ES.  Fig.~\ref{fig:qdobjective} shows results on the real mustard bottle experiments, where we see that while QD optimization improves the Medial Constraint performance, our method still significantly outperforms it. This demonstrates the value of $\relaxedtotalcost$ as a QD objective.

\section{Discussion and Future Work}
In sequential registration problems such as our probing experiments, we assume that the object is stationary and that the updated semantic points are given. However, keeping the object still while probing it is not trivial, as every contact has the potential to move the object. Rapid force and tactile feedback could minimize this issue. Contact could also be made with other objects during the probing motions. Contact point tracking and reasoning over object-contact associations is not within the scope of this paper. However, in future work we will explore using a method such as STUCCO \cite{zhong2022soft} to estimate object-contact associations and add the proper contacts to $\xall$. 

The experiments in this paper used a fixed sequence of probing motions. This makes for a fair comparison between methods, since the sequence is not dependent on any method's pose estimates. However, in practice, the next probing motion should depend on the current pose estimate. In future work we aim to explore how to reason over the plausible set of poses and plan trajectories that efficiently disambiguate between them, so as to localize the object with as few probing motions as possible.

\section{Conclusion}
We presented \methodabv{}, a pose registration method that utilizes point semantics, such as whether a point is in free space or on the object surface, to impose additional constraints and reduce pose ambiguity. 
Rather than a single best estimate, it produces a set of diverse plausible estimates given the observed data. We showed that it performs well on both simulated and real data collected from robot probing experiments. In particular, we separately demonstrated the value of performing Quality Diversity (QD) optimization for registration, and the strength of our proposed differentiable cost function as a QD objective. Additionally, we showed how to update the estimated transform set online with updated data, that \methodabv{} performs well on data with few contact points, and that it is seamless to integrate vision as an input modality.


\footnotesize
\bibliographystyle{plainnat}
\bibliography{references}

\newpage
\begin{appendices}
\section{CVO Performance}
\label{sec:cvoperf}
\begin{figure}[h]
    \centering
    \includegraphics[width=\linewidth]{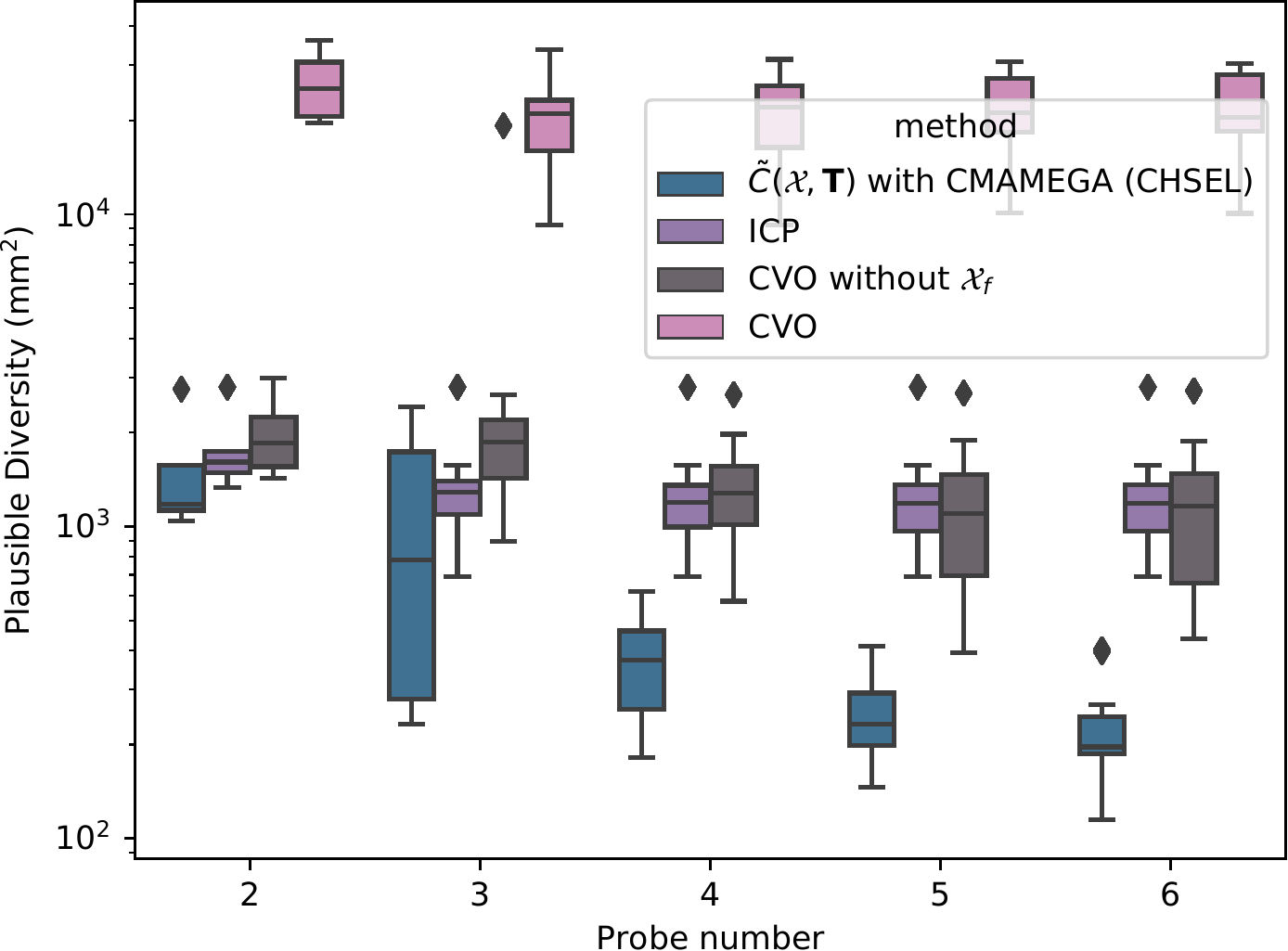}
    \caption{Plausible Diversity on the two real mustard bottle experiments with a focus on the improvement the CVO baseline receives from ignoring $\xfree$. Lower is better.}
    \label{fig:cvonofree}
\end{figure}

 Qualitatively, we noticed that CVO's estimated transforms tend to place the object such that $\xfree$ is in concentrated regions of observed free points. To check if this free/surface imbalance was the cause of CVO's poor performance, we ran CVO while ignoring $\xfree$ on the real mustard bottle experiments (results shown in Fig.~\ref{fig:cvonofree}). We see that CVO performs comparably to ICP when ignoring $\xfree$. This is significantly better than when considering $\xfree$, suggesting that CVO is not able to effectively use the free space information.

\end{appendices}

\end{document}